\documentclass[runningheads]{llncs}
\usepackage{graphicx}
\usepackage{cite}
\usepackage{amsmath,amssymb,amsfonts}
\usepackage{algorithmic}
\usepackage{graphicx}
\usepackage{textcomp}
\usepackage{xcolor}
\usepackage{multirow}
\usepackage{hyperref}
\usepackage{diagbox}
\usepackage{microtype}
\usepackage{booktabs}
\usepackage{xspace}
\usepackage{tabularx}
\usepackage{adjustbox}

\begin{document}

\def\degree{$^\circ$\xspace}

\title{ATSal: An Attention Based Architecture for Saliency Prediction in 360\degree Videos}
\titlerunning{ATSal: a saliency prediction model for 360\degree videos}
%
\author{Yasser Dahou\inst{1,}\thanks{Corresponding author}\and
Marouane Tliba\inst{2} \and
Kevin McGuinness\inst{1}\and
Noel O'Connor\inst{1}}
\authorrunning{Y. Dahou et al.}
%
\institute{Insight Centre for Data Analytics, Dublin City University, Dublin 9, Ireland
 \and
Institut National des Télécommunications et des TIC, Oran, Algeria\\ \email{yasser.dahoudjilali2@mail.dcu.ie}}
\maketitle

\begin{abstract}
The spherical domain representation of 360\degree video/image presents many challenges related to the storage, processing, transmission and rendering of omnidirectional videos (ODV). Models of human visual attention can be used so that only a single viewport is rendered at a time, which is important when developing systems that allow users to explore ODV with  head mounted displays (HMD). Accordingly, researchers have proposed various saliency models for 360\degree video/images. This paper proposes ATSal, a novel attention based (head-eye) saliency model for 360\degree videos. The attention mechanism explicitly encodes global static visual attention allowing expert models to focus on learning the saliency on local patches throughout consecutive frames. We compare the proposed approach to other state-of-the-art saliency models on two datasets: Salient360! and VR-EyeTracking. Experimental results on over 80 ODV videos (75K+ frames) show that the proposed method outperforms the existing state-of-the-art. 

\end{abstract}

\section{Introduction}
360\degree video, referred to as panoramic, spherical or omnidirectional video, is a recently introduced type of multimedia that provides the user with an immersive experience. The content of ODV is rendered to cover the entire $360\times180$ viewing space. Humans, however, naturally focus on the most attractive and interesting field-of-views (FoV) while ignoring others in their visual field, using a set of visual operations know as visual attention. Such selective mechanisms allow humans to interpret and analyze complex scenes in real time and devote their limited perceptual and cognitive resources to the most pertinent subsets of sensory data. Inspired by this visual perception phenomenon, saliency prediction or modeling is the process that aims to model the gaze fixation distribution patterns of humans on static and dynamic scenes. Modeling visual attention in omnidirectional video is an important component of being able to deliver optimal immersive experiences. The main objective is to predict the most probable viewports in a frame reflecting the average person's head movements (HM) and  eye movements (EM) reflecting the region-of-interest (RoI) inside the predicted viewport. Thus, when predicting the most salient pixels, it is necessary to predict both HM and EM for 360\degree video/image visual attention modeling.

Despite the remarkable advances in the field of visual attention modeling on a fixed viewport (see~\cite{b3,b4,b5} for a comprehensive review), saliency prediction studies on 360\degree video/image are still limited. This is in part due to the comparatively small amount of research that has investigated the visual attention features that affect human perception in panoramic scenes. This is further compounded by the lack of commonly used large-scale head and eye-gaze datasets for 360\degree content as well as the associated difficulties of using this data compared with publicly available 2D stimuli datasets. 

A recent survey conducted by Xu et al.~\cite{b1} reviewed many works for predicting the HM/EM saliency maps of 360\degree images and video that model the probability distribution of viewports/RoIs of multiple subjects. Rai et al.~\cite{b17} derived the volcano-like distribution of EM with the viewport. Results show that eye-gaze fixations are quasi-isotropically distributed in orientation, typically far away from the center of the viewport. Furthermore, ODV presents some statistical biases, as investigated in~\cite{b18}. Human attention on 360\degree video is biased toward the equator and the front region, known as the equator bias, which could be leveraged as priors in modelling. Along with the statistical bias, a subject's attention is driven by the most salient objects in the scene. It has been shown that a smaller number of closer objects capture more human attention~\cite{b19}. 

Motivated by this, our novel approach combines global visual features over the entire view with features derived from local patches in consecutive frames. A good set of features are those that share the minimal information necessary to perform well at the saliency task. The goal of our representation is to capture as much information as possible about the stimulus. This technique results in state-of-the-art accuracy on standard benchmarks.

The contributions of this paper are as follows:
\begin{itemize}
    \item We demonstrate the importance of global visual attention features in achieving better saliency performance.
    \item A new approach, ATSal, is presented that combines an attention mechanism with expert instances for each patch location to learn effective features for saliency modelling.
    \item We pre-process the VR-EyeTracking~\cite{b6} dataset by extracting the well annotated fixation/saliency maps from the provided raw data and make this available to others to use at this \textcolor{blue}{\href{https://drive.google.com/drive/folders/1y6hCYOdn7BbBguxfBfOhgoUW_1DygET1?usp=sharing}{link}}.
    \item We compare our approach against a representative selection of state-of-the-art 360\degree approaches on both the VR-EyeTracking and Salient360! datasets.
\end{itemize}

The rest of the paper is organized as follows: Section 2 provides an overview of related 360\degree video saliency works. Section 3 gives a detailed description of the proposed framework. Section 4 compares the experimental results to state-of-the-art methods. Finally, we conclude this work in Section 5. The results can be reproduced with the source code and trained models available on GitHub: \textcolor{blue}{\href{https://github.com/mtliba/ATSal}{link}}.

\section{Related work}

In this section, we present the important works related to attention  modelling  for 2D dynamic stimuli and 360\degree  video/image. They  mainly  refer  to  predicting  the HM/EM saliency maps of 360\degree video/images and can be further grouped into heuristic approaches and data-driven approaches.

\subsection{2D dynamic saliency models}
Video saliency prediction has advanced significantly since the advent of deep learning. Recently, many works have investigated deep neural networks (DNNs) for video saliency prediction  (e.g.~\cite{b7,b8,b9,b10,b11,b12,b13}). SalEMA~\cite{b10} added a conceptually simple exponential moving average of an internal convolutional state to the SalGAN network~\cite{b14}. 3DSAL~\cite{b11} performs 3D convolutions on the generated VGG-16 features, while using a weighting cuboid function to smooth the spatial features; the novel contribution is learning saliency by fusing spatio-temporal features. Unisal~\cite{b7} proposed four novel domain adaptation techniques to enable strong shared features: domain-adaptive priors, domain-adaptive fusion, domain-adaptive smoothing, and Bypass-RNN. Unisal achieved the state-of-the-art results on the DHF1K benchmark~\cite{b12} (e.g. AUC-J:~0.901, NSS:~2.776).

\subsection{360\degree Heuristic approaches}
The heuristic approaches encode saliency on the 360\degree sphere using handcrafted features. The pioneer works by Iva et al.~\cite{b15,b16} generate the spherical static saliency map by combining together chromatic, intensity, and three cue conspicuity maps after normalization, through multiscale analysis on the sphere. They build the motion pyramid on the sphere by applying block matching and varying the block size. Finally, the two maps are fused to produce the dynamic saliency map on the sphere. Unfortunately, no quantitative results were provided in~\cite{b15,b16} since the HMD was not yet available at the time. RM3~\cite{b26} for 360\degree images, combines low-level features hue, saturation and GBVS features, with the high-level features present in each viewport. Unlike~\cite{b15,b16,b26}, Fang et al.~\cite{b27} proposed the extraction of low-level features of color, texture luminance, and boundary connectivity from the super-pixels at multiple levels segmented from the input equirectangular projection (ERP) image.

Other works have adapted existing 2D saliency models into 360\degree video/images. This approach, however, suffers from geometric distortion and border artifacts. Abreu et al.~\cite{b20} introduced a fused saliency map (FSM) approach to HM saliency prediction on 360\degree images; they adapted SALICON~\cite{b21} (a 2D image saliency prediction model) to 360\degree images using ERP. Lebreton et al.~\cite{b22} extended the 2D Boolean Map Saliency  (BMS~\cite{b23}) and Graph-Based Visual Saliency (GBVS~\cite{b24}) models to integrate the properties for equirectangular images, naming their approaches BMS360 and GBVS360 respectively. Maugey et al.~\cite{b25} applied a 2D saliency model on each face generated under the cubemap projection (CMP). Recently, it has become easier to collect HM and EM data and thus there have emerged many end-to-end saliency prediction approaches for 360\degree video/images.

\subsection{360\degree Data-driven approaches}
Trained on recently published datasets~\cite{b18,b6,b28,b29}, a number of deep neural netrowk (DNN) saliency prediction approaches for 360\degree video have been proposed~\cite{b29,b18,b30,b31,b32,b33,b34,b35}. Nguyen et al.~\cite{b33} fine-tuned the PanoSalNet 2D static model on 360\degree video datasets to predict HM saliency map of each frame without considering the temporal dimension. The predicted saliency is enhanced by a prior filter based on an a-priori statistical bias. Cheng et al.~\cite{b30} proposed a DNN-based spatial-temporal network, consisting of a static model and a ConvLSTM module to adjust the outputs of the static model based on temporal features. They also aggregated a ``Cube Padding'' technique in the convolution, pooling, and convolutional LSTM layers to keep connectivity between the cube faces by propagating the shared information across the views. Lebreton et al.~\cite{b32} extended BMS360~\cite{b22} to V-BMS360 by adding a temporal bias and optical flow-based motion features. Hu at al.~\cite{b35} proposed a deep learning-based agent for automatic piloting through 360\degree sports videos. At each frame, the agent observes a panoramic image and has the knowledge of previously selected viewing angles. The developed online policy allows shifting the current viewing angle to the next preferred one through a recurrent neural network. Fang et al.~\cite{b36} fine tuned SalGAN~\cite{b4} on the Salient360!~image dataset with a new loss function combining three saliency metrics. Qiao et al.~\cite{b37} proposed a Multi-Task Deep Neural Network (MT-DNN) model for head movement prediction; the center of each viewport is spatio-temporally aligned with 8 shared convolution layers to predict saliency features.

Unlike all previous approaches,  Zhang et al.~\cite{b29} proposed a spherical convolutional neural network, spherical U-NET, trained following the teacher forcing technique, for applying a planar saliency CNN to a 3D geometry space, where the kernel is defined on a spherical crown, and the convolution involves the rotation of the kernel along the sphere. The model input includes one frame and the predicted saliency maps of several previous frames to allow for improved modelling of dynamic saliency. The authors also defined a spherical MSE (S-MSE) loss function for training the spherical U-Net to reduce the non-uniform sampling of the ERP. Furthermore, instead of using supervised learning to learn saliency from data, Mai at al.~\cite{b18} applied deep reinforcement learning (DRL) to predict HM positions by maximizing the reward of imitating human HM scanpaths through the agent's actions. The reward, which measures the similarity between the predicted and ground-truth HM scanpaths, is estimated to evaluate the action made by the DRL model. The reward is then used to make a decision on the action through the DRL model: i.e., the HM scanpath in the current frame. 

It is clear from the above review that the number of models targeting  HM and EM 360\degree video visual attention modeling is still considerably limited compared with 2D video saliency prediction. There is still much work to be done to meet the specific requirements of ODV. The contributions of this paper as outlined above attempt to address this. 

\section{Proposed model}
\begin{figure*}[t]
\makebox[\linewidth]{
\includegraphics[scale=0.33]{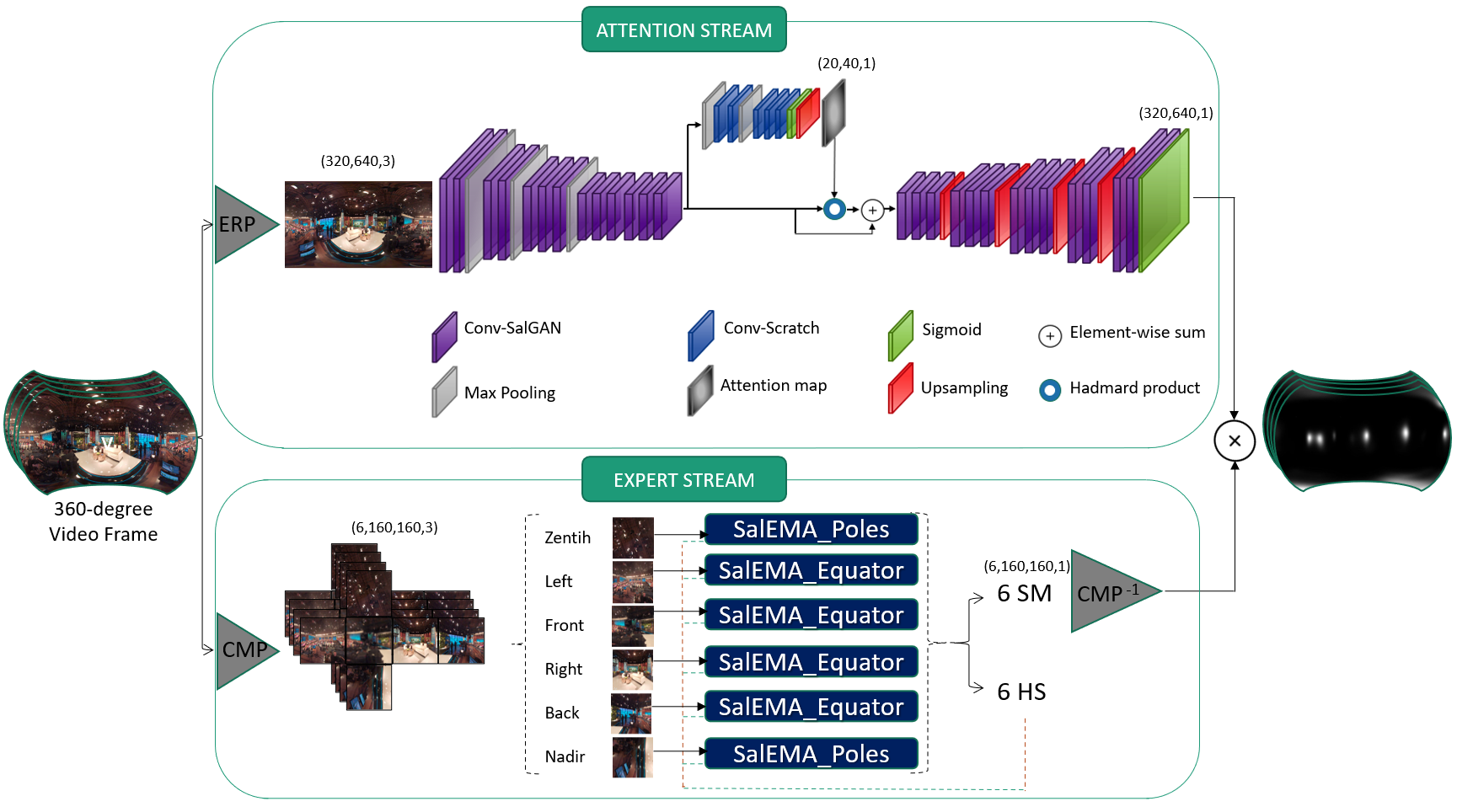}
}
\caption{Network Architecture of ATSal. The attention stream is for extracting global saliency. The experts model operates on local patches. The final saliency is obtained by pixel-wise multiplication between the two maps generated by each stream.} 

\label{fig1}
\end{figure*}

This section describes our framework (see Figure.~\ref{fig1}). The design consists of an attention model, expert models, and the fusion method.

Our approach operates in two parallel streams. One stream is dedicated to extracting global attention statistics via the attention mechanism applied on the ERP frame. This explicitly captures static saliency information allowing the second steam to learn more effective temporal features. Moreover, the expert models embed two instances of SalEMA on the cube faces, \textbf{SalEMA\_Poles} for the Zenith and Nadir and \textbf{SalEMA\_Equator} specialized for the equator viewports. This is motivated by findings that investigated the properties of  panoramic scenes that argued that the fixations distribution is highly correlated with the locations of the viewports~\cite{b37}. Thus, we adapted SalEMA weights to meet these requirements.

\subsection{Attention mechanism}
Despite the strong research interest and  investigations of saliency prediction over 360\degree stimuli, to the best of our knowledge, no previous work has exploited the global attention effect related to omnidirectional scenes in general and for head and eye saliency prediction in particular. We address this by implementing an attention mechanism in a parallel stream which is able to encode the global statistics of the 360\degree video/image. A dense mask is learned at the middle of the bottleneck with an enlarged receptive field, and this attention mask is combined by performing pixel-wise multiplication with the feature map followed by pixel-wise summation among the channels to obtain the final optimal latent variable representation, thereby embedding as much information as possible about the input frame.

\textbf{Why an attention mechanism?} Recent studies~\cite{b38,b39,b40,b41} have shown the effectiveness of attention mechanisms for several computer vision downstream tasks. In such approaches, attention is learned in an adaptive, joint, and task-oriented manner, allowing the network to focus on the most relevant parts of the input space. Similar to~\cite{b12}, our latent attention module extracts strong spatial saliency information and was trained on both Salient360!~\cite{b42} and Sitzman~\cite{b43} 360 image saliency datasets in a supervised fashion. Such a design leads to generalization of the prediction performance across different 360 saliency datasets. As shown in Fig.\ref{archi}, the input is an ERP image $X\in  \mathbb{R}^{640\times320\times3}$. This shape implies the preservation of the initial spherical characteristics of the data distribution, which is of ratio 2:1. However, a larger receptive field is needed to propagate each pixel-wise response to a field of $640 \times 320$ which is the sufficient condition of having the global attention. The approach is motivated by the following considerations:

\begin{table}
\caption{Attention model results on Salient360! validation images.}
\centering
\begin{tabular}{p{3.5cm} p{1.3cm}p{1.3cm}p{1.3cm}p{1.3cm}p{1.3cm}}
\toprule
      {} &
      \multicolumn{5}{c}{Salient360!}\\
     
 &AUC-J $\uparrow$ &NSS $\uparrow$ &CC $\uparrow$ &SIM $\uparrow$ &KLD $\downarrow$ \\
      
\midrule
Attention model  &0.837&1.680& 0.631&0.629&0.801\\

\bottomrule
\end{tabular}
\label{resattention}
\end{table}

\begin{itemize}
    \item As investigated in~\cite{b44}, human attention is driven by complementary pathways, which compete on global versus local features. Global features such as: semantics, pose, and spatial configuration, called scene layout or ``gist'', guides the early visual processing. In a panoramic scenario, the CMP prediction forces the model to lose the global contextual information while considering each face separately. Through the attention mechanism, the auto-encoder is forced to learn a more explicit global static representation. The global and local features are disentangled for robust predictions.
    
    \item  The receptive field of the encoder needs to cover the entire input. To this end, we modified the VGG-16 encoder, by deleting the last two max pooling layers to better preserve  spatial frequencies, and set the max poling layer-3 stride factor and kernel size from (2 $\to$ 4). This replaces the initial VGG-16 receptive field ($212\times212$) with an enlarged one of ($244\times244$) but this is still not sufficient to cover the whole input $X\in  \mathbb{R}^{640\times320\times3}$. To tackle this issue, we extended the $conv5$ layer with an attention mechanism containing several convolution layers interspersed with pooling and up-sampling operations. Given the latent variable:
    \begin{equation}
z_{1}=f_{\theta}(X) \in \mathbb{R}^{20\times40\times512},
\end{equation}
where $f_{\theta}$ is the encoder, the attention mechanism yields a dense attention map $M \in \big[0,1]^{20\times40}$ with an enlarged receptive field of $676\times676$.

\end{itemize}

The attention module is activated with a sigmoid function, which relaxes the sum to one constraint often used in softmax based neural attention.  The feature map $z_{1}$ is further enhanced by:
\begin{equation}
z^{c}= (1+M)\circ z_{1}^{c},
\end{equation}
where $c \in \{1,...,512\}$ is the channel index and $M$ operates as a feature wise selector on $z_{1}$, with the residual connection to keep all the useful information~\cite{b45}.  The optimal latent variable $z \in \mathbb{R}^{20\times40\times512}$ is fed to the decoder to learn the predicted saliency map $Y_{1} \in \mathbb{R}^{640\times320}$. Table.~\ref{archi} summarizes the architecture of the attention mechanism.

\begin{table}
\caption{\label{archi}Architecture of the attention mechanism}

\centering

\begin{tabular}{p{4cm} p{4cm} p{3cm}}

\toprule
Layer (type) & Output shape & \# Parameters\\

\midrule

MaxPool2D\_1 & $1\times 512 \times 10 \times 20$ & 0\\

Conv2D\_1 &$1\times 64\times 10\times 20$ & 294,976\\

ReLU &$1\times 64\times 10\times 20$ & 0\\

Conv2D\_2 & $1\times 128\times 10\times 20$ & 73,856\\

ReLU &$1\times 128\times 10\times 20$ & 0\\

MaxPool2D\_2 &$1\times 128\times 5\times 10$ & 0\\

Conv2D\_3 &$1\times 64\times 5\times 10$ & 73,792\\

ReLU &$1\times 64\times 5\times 10$ & 0\\

Conv2D\_4 &$1\times 128\times 5\times 10$ & 73,856\\

ReLU &$1\times 128\times 5\times 10$ & 0\\

Conv2D\_5 &$1\times 1\times 5\times 10$ & 129\\

Upsample &$1\times 1\times 20\times 40$&0 \\

Sigmoid &$1\times 1\times 20\times 40$ & 0\\

\midrule
 \multicolumn{3}{c}{Total Parameters: 516,609} \\
\bottomrule
\end{tabular}

\end{table}

\subsection{Expert models}

All previous works learn the same network parameters for the six faces of the cube. However, the saliency density is mostly represented in the equator, forcing the network to over estimate fixations in the poles, which is one of the main reasons the prediction performance drops. In short, the expert stream of our framework instantiates two expert versions from SalEMA~\cite{b10} and combines their results through the inverse projection to predict the final saliency. The main point of the expert stream is to predict 360\degree  dynamic saliency viewports based on both spatiotemporal content and viewport location. Accordingly, each viewport is predicted independently with shared weights.  
The input frame in ERP format $X$ is first projected to the CMP format resulting in $x_{i}, i \in \{1,..6\}$. Where $x_{0,1,2,3}$ represents the front, left, right and back views, and $x_{4,5}$ are the Zenith and Nadir views. The original SalEMA was trained on DHF1K dataset using the binary cross entropy (BCE) loss and a SalGAN encoder-decoder with an added exponential moving average (EMA) recurrence. The exponential weighted average takes the state $S_{t}$ output by a convolutional layer at time $t$. The output $E_{t}$ is then propagated the next layers using a convex combination of the state $S_t$ and previous states:
 \begin{equation}
E_{t}= \alpha S_{t}+(1-\alpha)E_{t-1}.
\end{equation}

with initial $\alpha=0.1$. We further changed the training protocol of SalEMA, by adjusting each corresponding cube face into a batch of 80 for the poles, but we kept the initial 10 for the equators; also, we kept BCE as the objective function for fine-tuning SalEMA. This is motivated by the low motion present in each viewport (see~\cite{b46}). \textbf{SalEMA\_Poles} was fine-tuned on batches of the faces $x_{4,5}$ to capture the underlying features on these viewports, while \textbf{SalEMA\_Poles} was adapted to faces $x_{0,1,2,3}$. The generated local saliency per cube face is then inversely projected into the $ERP$ format using $CMP^{-1}$; we denote $Y_{2} \in \mathbb{R}^{640\times320}$ the generated saliency map after the inverse projection. The final saliency map $Y_{t}$ is obtained after pixel wise multiplication between $Y_{1}$ and $Y_{2}$:
\begin{equation}
Y= Y_{1} \circ Y_{2} \in  \big[0,1]^{640\times320}
\label{eq4}
\end{equation}

\subsection{Loss function for the attention stream}

According to~\cite{b47}, the saliency metrics cover different aspects of the saliency map. Thus, we define the loss function as a combination between the Kullback-Leibler Divergence (KL) and the Normalized Scanpath Saliency (NSS). We denote the predicted saliency $Y_{1} \in \big[0,1]^{640\times320}$, the fixation map as $F \in \{0,1\}^{640\times320}$, the dense mask at the middle of the bottleneck $M \in \big[0,1]^{20\times40}$, and the continuous saliency map obtained after blurring $F$ with a Gaussian filter ($\sigma=9.35^\circ)$ as  $Q_{1} \in \big[0,1]^{640\times320}$. $Q_{2} \in \big[0,1]^{40\times20}$ is the down-sampled version. The attention stream loss function is defined as follows:
\begin{equation}
\begin{split}
\mathcal{L}_\mathcal{T} (Y_{1},M,F,Q_{1},Q_{2}) = \alpha_{1}\mathcal{L}_\text{KL}(Y_{1},Q_{1}) + \\ 
\alpha_{2}\mathcal{L}_\text{NSS}(Y_{1},F) + \beta\mathcal{L}_{\text{KL}}(M,Q_{2})
\end{split}  
\end{equation}
where $\alpha_{2}=\beta=0.2$, and $\alpha_{2}=0.8$. ${\cal{L}}_{KL}$ is chosen as the primary loss:
 \begin{equation}
 \label{eq5}
\mathcal{L}_{\text{KL}}(Y_{1},Q_{1}) = \sum_i {Q_{1}}_{i}\log\left(\epsilon+\frac{{Q_{1}}_{i}}{\epsilon+{Y_{1}}_{i}}\right).
\end{equation}
${\mathcal{L}}_{\text{NSS}}$ is derived from the NSS metric, which is a similarity metric. We therefore optimize its negative:
 \begin{equation}
 \label{eq20}
{\cal{L}}_{NSS}(Y_{1},F) = -\frac{1}{N}\sum_i \bar{Y}_{1i} \times F_{i},
\end{equation}
where $N=\sum F_{i}$ and $\bar{Y}_{1i} = (Y_{1i} -\mu(Y_{1i})) / \sigma(Y_{1i}).$

\section{Experiments}

\begin{figure*}[t]
\makebox[\textwidth]{
\includegraphics[scale=0.21]{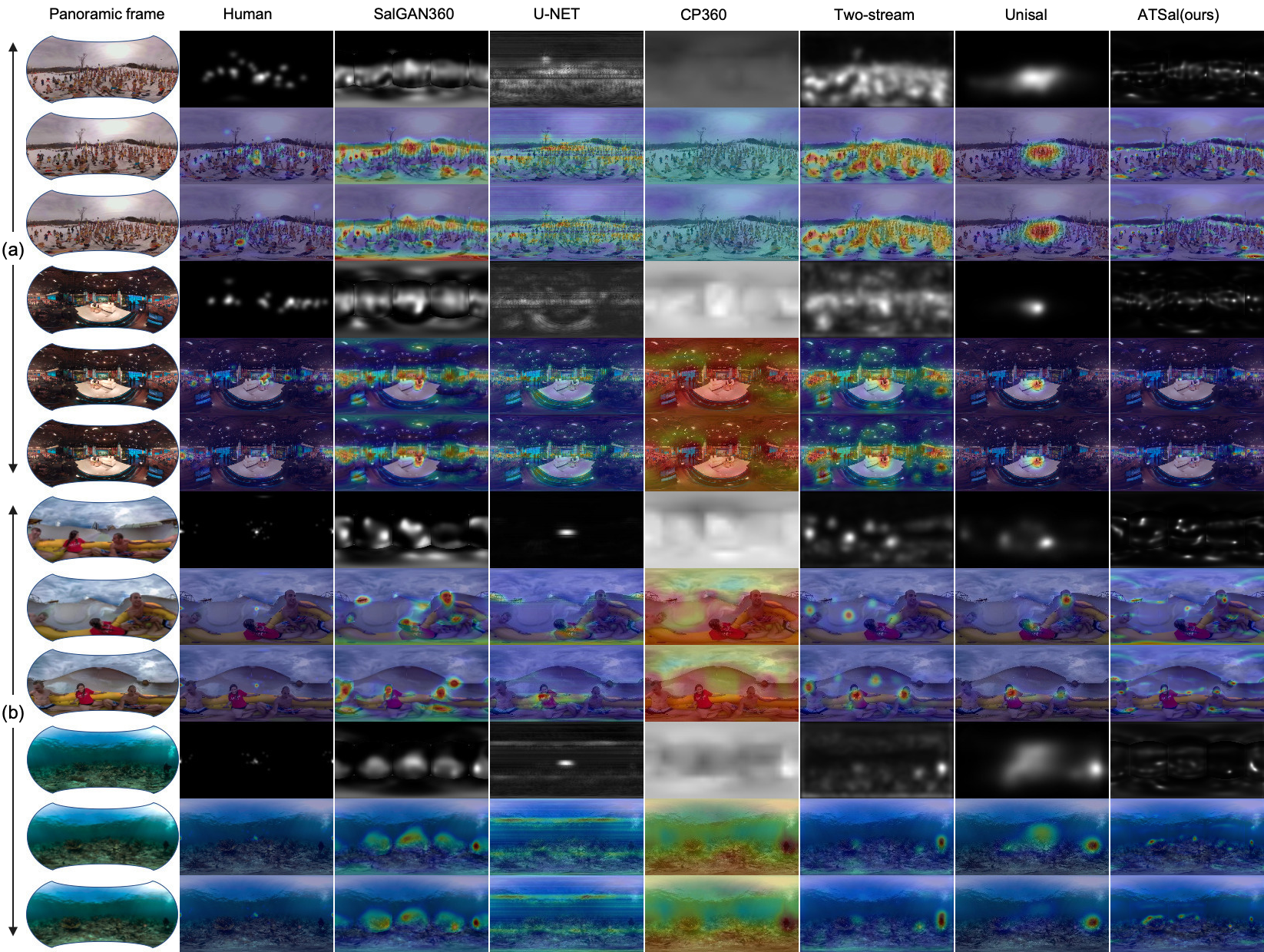}
}
\caption{Qualitative results of our ATSal model and five other competitors on sample frames from  VR-EyeTracking and Salient360! datasets. It can be observed that the proposed ATSal is able to handle various challenging scenes well and produces consistent omnidirectional video saliency results.}
\label{fig2}
\end{figure*}

\begin{figure*}
\centering
\makebox[\linewidth]{
\includegraphics[scale=0.112]{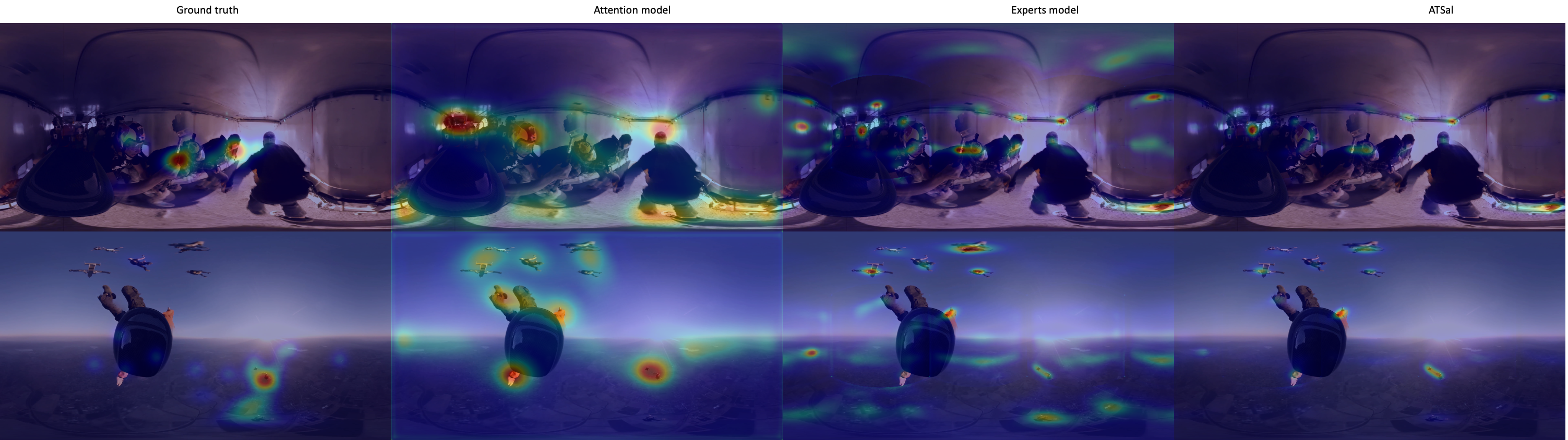}
}
\caption{Overlaid saliency of the attention model, experts model, and the final ATSal. It can be seen that the attention model captures more global saliency, while the experts focus on local patches. ATSal gathers both features for the final prediction.}
\label{fig3}
\end{figure*}

\subsection{Experimental setup}
\textbf{Training.} The attention model was trained in two stages. First, to encode static saliency, we trained the attention model on  360\degree images from the Salient360!~and Sitzman datasets. Due to the small amount of labelled static data (103 omnidirectional images), we applied some common data augmentation techniques: mirroring, rotations, and horizontal flipping. The resulting dataset contains 2,368 360\degree images. Results on the images Salient360!~validation set are shown in Table.~\ref{resattention}. For dynamic stimuli, we consider 2 settings: training both streams on: (i) Video-Salient360!, (ii) VR-EyeTracking. For Salient360!, we used 15 ODV for training and 04 for validation. For VR-EyeTracking, the training and testing sets include 140 and 75 videos respectively. We evaluate our model on the the test set of VR-EyeTracking and the validation sets (images and videos) of Salient360! (due to the unavailability of the  reserved set after we contacted the authors), in total 79 ODVs with over 75,000 frames.

\textbf{Competitors.} ATSal is compared with seven models corresponding to three state-of-the-art 2D video saliency models: Unisal~\cite{b7}, 3DSAL~\cite{b11}, SalEMA~\cite{b10}, and four 360\degree specialized models: U-NET~\cite{b29}, CP360~\cite{b30}, SalGAN360~\cite{b36}, and Two-Stream~\cite{b34}. This choice is motivated by the publicly available code. However, U-NET inference protocol was not discussed in the paper since the authors use the teacher forcing technique to train the model, so we feed the  predicted saliency maps at time $t-1$ into the model at  time $t$. 
All models were evaluated according to five different saliency metrics: Normalized Scanpath Saliency (NSS), Kullback-Leibler Divergence(KLD), Similarity (SIM), Linear Correlation Coefficient (CC), and AUC-Judd (AUC-J). Please refer to~\cite{b47} for an extensive review of these metrics.

\textbf{Technical details}. The attention model is implemented in Pytorch. The Adam optimizer was used for the training with a learning rate of $10^{-5}$. The attention model was trained end-to-end using a NVIDIA GTX 1080 and a 3.90 GHz I7 7820 HK Intel processor. SalEMA experts were fine-tuned with a modified input size of $160\times160$.

\textbf{Dataset processing}. The VR-EyeTracking dataset recorded by~\cite{b47}, consists of 208 diverse content 4k ODVs, where the head/eye saccades were obtained from 45 participants. However, neither fixation nor saliency maps were published publicly. We processed the gaze recording to obtain the $2048\times1024$ fixation maps. Saliency maps were generated by convolving each fixation point (for all observers of one video) with a Gaussian filter, with $\sigma=9.35^\circ$ for head and eye data. The processed dataset is now available at this \textcolor{blue}{\href{https://drive.google.com/drive/folders/1y6hCYOdn7BbBguxfBfOhgoUW_1DygET1?usp=sharing}{link}}. 

\subsection{Results}
Table.~\ref{results} shows the comparative study with the aforementioned models according to the different saliency metrics on Salient360!~and VR-EyeTracking datasets (4/75) test ODVs. Our model is very competitive in the two datasets: ATSal exhibits  the  best  score for all metrics. Surprisingly, 2D approaches achieve mostly the same results as 360\degree specialized models; this points to the potential to consider the direct transfer of the well verified visual attention features from 2D to 360\degree, but also perhaps reflects the lack of large scale well-annotated 360\degree datasets. We invite the reader to watch the qualitative results of the proposed method in the form of a video available \textcolor{blue}{\href{https://youtu.be/-pTILQYMD9w}{here}}. 

\begin{table*}[t]
\caption{Comparative performance study on: Salient360!~\cite{b28}, VR-EyeTracking~\cite{b6} datasets.}
\centering

\begin{adjustbox}{max width=\textwidth}

\begin{tabular}{ll|rrrrr|rrrrr}
\toprule
      {}&{\diagbox{Models}{Dataset}} &
      \multicolumn{5}{c|}{Salient360!} &
      \multicolumn{5}{c}{VR-EyeTracking}\\
     
      &&AUC-J $\uparrow$ &NSS $\uparrow$ &CC $\uparrow$ &SIM $\uparrow$ &KLD $\downarrow$ &AUC-J $\uparrow$ &NSS $\uparrow$ &CC $\uparrow$ &SIM $\uparrow$ &KLD $\downarrow$ \\
      
\midrule
\multirow{3}{*}{2D models}
&Unisal~\cite{b7}  &0.746&1.258&0.206&0.190&8.524&0.764&1.798&0.314&0.281&6.767\\
 
&SalEMA~\cite{b10} &--&--&--&--&--&0.772&1.790&0.320&0.284&6.532 \\

&3DSal~\cite{b11} &0.656&0.898&0.192&0.139&10.129&0.679&1.229&0.228&0.232&8.317\\
\midrule
\multirow{4}{*}{360\degree models}
 &U-NET ~\cite{b29}
&0.725&0.864&0.129&0.122&9.810&0.818&1.430&0331&0.247&7.070 \\
 &Cheng et al.~\cite{b30} &0.819&1.094&0.175&0.213&9.313&0.735&0.914&0.171&0.179&8.879 \\
&Salgan360~\cite{b36} &--&--&--&--&--&0.704&1.267&0.236&0.238&7.625 \\
&Two-Stream~\cite{b34}  &0.787&1.608&0.265&0.179&8.252&0.827&1.906&0.346&0.254&7.127\\
\midrule

\multirow{3}{*}{Training setting (i)}
&Attention   & 0.827& 1.397& 0.222&0.168& 8.225&0.795 &1.338& 0.255& 0.229& 7.405 \\

&Experts  & 0.870 &2.438& 0.366& 0.227 & 6.990& 0.754& 1.292& 0.221& 0.224& 7.840 \\

&Ours & \textbf{0.881}&\textbf{2.580}&\textbf{0.363}&\textbf{0.252}&\textbf{6.327}&0.801&1.590&0.265&0.255&6.571\\
\midrule
\multirow{3}{*}{Training setting (ii)}
&Attention   & 0.796&1.299&0.201&0.168&7.973&0.836&1.791&0.342&0.302&5.664\\
 
&Experts &0.807&1.564&0.242&0.223&7.679&0.778&1.388&0.259&0.211&7.407 \\

&Ours  &0.837&1.764&0.285&0.255&7.382&\textbf{0.862}&\textbf{2.185}&\textbf{0.363}&\textbf{0.312}&\textbf{5.561} \\
\bottomrule
\end{tabular}
\end{adjustbox}
\label{results}
\end{table*}

\subsection{Performance study}

{\bf{VR-EyeTracking}}. The 75 diverse test ODVs of this dataset makes the prediction task very challenging. ATSal outperforms all competitors according to all five metrics due to the combination of global features with the attention stream, but also due to the local patches encoding the temporal domain predicted by the experts model. 

{\bf{Salient360!}}. On both distributions of the Salient360! dataset (images and videos), the proposed model gains a substantial quantitative advantage in accuracy compared with other models. This demonstrates the capacity of our model to handle a larger set of scenarios.

Encouraged by the positive results of the attention model trained on static images, and later fine-tuned for the Salient360! and VR-EyeTracking ODVs.
We evaluated the model components separately on both datasets. Table.~\ref{results} indicates the competitiveness of the attention model with respect to other competitors, even without the integration of the expert model. A possible explanation could relate to smaller motion effects in ODVs compared to 2D videos, where viewport locations are the most prominent features. 

Figure.~\ref{fig2} illustrates the prediction task on a sample of 360\degree frames from the two datasets: Salient360!~and VR-EyeTracking. It can be seen that the generated saliency maps with ATSal are more comprehensive and look remarkably similar to the Ground truth maps in terms of saliency distribution. Other competitors shown in the same figure overestimate saliency in general, or overlay bias the equator/center. Furthermore, the effectiveness of ATSal in capturing the main objects in the scene is clear.

For more qualitative results, Figure.~\ref{fig3} shows the overlaid saliency maps on sample frames from Salient360! for the attention model, expert models, and ATSal. Two key points can be seen from these figures:

\begin{itemize}
 \item
    Figure.~\ref{fig2}: ATSal predicts consistent saliency at the very beginning of each video due to the attention mechanism modeling the spatial features, while other competitors center the saliency around the equator. As the scene progresses, ATSal ignores some static regions and only focuses on other moving viewports. In video (b), all models only detect faces near the equator, while ATSal also attends to other parts of the scene. 
    \item 
    Figure.~\ref{fig3}: the generated saliency maps using only the attention model are sparse and cover all the possible outcomes in the scene. This is due to the enlarged receptive field of the latent space: the model tends to give a high probability to a given pixel, which makes it salient. The expert models act as a saliency selector through the pixel wise multiplication. This approach encourages not overestimating saliency, but also forces the model to generate more focused and consistent saliency regions.
    \end{itemize}

\begin{table}
\caption{GPU inference time comparison of video saliency prediction methods (NVIDIA GTX 1080). All methods are reported based on the VR-EyeTracking benchmark~\cite{b6}. Best computational performance among dedicated 360\degree  models is shown in bold.}
\centering
\begin{tabular}{p{6cm}p{1.7cm}}

\toprule
     Model&Runtime (s) \\
\midrule
SalGAN360~\cite{b36} & 17.330\\
     Two-stream ~\cite{b34} &2.061 \\
     U-Net~\cite{b29} & 1.802\\
     (*) 3DSAL~\cite{b11} & 0.100\\
    (*) SalEMA~\cite{b10} & 0.020\\
     (*) Unisal ~\cite{b7} & 0.010\\

     \midrule
     Attention (ours)& 0.050\\

     Experts (ours)& 0.170\\
     
     \textbf{ATSal (ours)}& \textbf{0.230}\\
\bottomrule
\multicolumn{2}{c}{(*) 2D models.}\\
\end{tabular}
\label{runtime}
\end{table}
\textbf{Computational load.} Model efficiency is a key factor for real-time 360\degree videos application like streaming. Table.~\ref{runtime} shows a GPU runtime comparison (processing time per 360\degree frame) of the different competitors on the 4K VR-EyeTracking ODVs. Compared with other 360\degree specialized models, ATSal is over $9\times$ faster than Two-stream, which is the current state-of-the-art model according to the Salient360! leaderboard.

\section{Conclusion}
We proposed a novel deep learning based 360\degree video saliency model that embeds two parallel streams encoding both the global and local features. The attention model explicitly captures the global visual static saliency information through an attention mechanism at the middle of the bottleneck with an enlarged receptive field propagating the contextual pixel-wise information to the whole frame space. This design allows the expert models to learn more effective local region based saliency. The temporal domain is augmented using the simple exponential moving average (EMA).

We performed extensive evaluations on the Salient360! and VR-EyeTracking datasets, and compared the results of our model with the previous 2D and 360\degree static and dynamic models. The qualitative and quantitative results have shown that the proposed method is consistent, efficient, and outperforms other state-of-the-art.

\section*{Acknowledgement}
This publication has emanated from research supported by Science Foundation Ireland (SFI) under Grant Number SFI/12/RC/2289\_P2, co-funded by the European Regional Development Fund, through the SFI Centre for Research Training in Machine Learning (18/CRT/6183).


\begin{thebibliography}{00}
\bibitem{b1}
Xu, M., Li, C., Zhang, S., \& Le Callet, P. (2020). State-of-the-art in 360 video/image processing: Perception, assessment and compression. IEEE Journal of Selected Topics in Signal Processing, 14(1), 5-26.

\bibitem{b2}
De Abreu, A., Ozcinar, C., \& Smolic, A. (2017, May). Look around you: saliency maps for omnidirectional images in VR applications. In 2017 Ninth International Conference on Quality of Multimedia Experience (QoMEX) (pp. 1-6). IEEE.

\bibitem{b3} 
Itti, L.,  Koch, C. (2000). A saliency-based search mechanism for overt and covert shifts of visual attention. Vision research, 40(10-12), 1489-1506.

\bibitem{b4} 
Pan, J., Ferrer, C. C., McGuinness, K., O'Connor, N. E., Torres, J., Sayrol, E., Giro-i-Nieto, X. (2017). SalGAN: Visual saliency prediction with generative adversarial networks. arXiv preprint arXiv:1701.01081.

\bibitem{b5}
Borji, A. (2018). Saliency prediction in the deep learning era: An empirical investigation. arXiv preprint arXiv:1810.03716, 10.

\bibitem{b6}
Xu, Y., Dong, Y., Wu, J., Sun, Z., Shi, Z., Yu, J.,  Gao, S. (2018). Gaze prediction in dynamic 360 immersive videos. In proceedings of the IEEE Conference on Computer Vision and Pattern Recognition (pp. 5333-5342).

\bibitem{b7}
Droste, R., Jiao, J.,  Noble, J. A. (2020). Unified Image and Video Saliency Modeling. arXiv preprint arXiv:2003.05477.

\bibitem{b8}
Min, K.,  Corso, J. J. (2019). TASED-net: Temporally-aggregating spatial encoder-decoder network for video saliency detection. In Proceedings of the IEEE International Conference on Computer Vision (pp. 2394-2403).
ISO 690	

\bibitem{b9}
Lai, Q., Wang, W., Sun, H.,  Shen, J. (2019). Video saliency prediction using spatiotemporal residual attentive networks. IEEE Transactions on Image Processing, 29, 1113-1126.

\bibitem{b10}
Linardos, P., Mohedano, E., Nieto, J. J., O'Connor, N. E., Giro-i-Nieto, X.,  McGuinness, K. (2019). Simple vs complex temporal recurrences for video saliency prediction. British Machine Vision Conference (BMVC), 2019

\bibitem{b11}
Djilali, Y. A. D., Sayah, M., McGuinness, K.,  O'Connor, N. E. (2020). 3DSAL: an efficient 3D-CNN architecture for video saliency prediction.

\bibitem{b12}
Wang, W., Shen, J., Guo, F., Cheng, M. M., Borji, A. (2018). Revisiting video saliency: A large-scale benchmark and a new model. In Proceedings of the IEEE Conference on Computer Vision and Pattern Recognition (pp. 4894-4903).

\bibitem{b13}
Bak, C., Kocak, A., Erdem, E., Erdem, A. (2017). Spatio-temporal saliency networks for dynamic saliency prediction. IEEE Transactions on Multimedia, 20(7), 1688-1698.

\bibitem{b14}
Pan, J., Sayrol, E., Nieto, X. G. I., Ferrer, C. C., Torres, J., McGuinness, K.,  OConnor, N. E. (2017, July). Salgan: Visual saliency prediction with adversarial networks. In CVPR Scene Understanding Workshop (SUNw).

\bibitem{b15}
Bogdanova, I., Bur, A., Hügli, H.,  Farine, P. A. (2010). Dynamic visual attention on the sphere. Computer Vision and Image Understanding, 114(1), 100-110.

\bibitem{b16}
Bogdanova, I., Bur, A.,  Hugli, H. (2008). Visual attention on the sphere. IEEE Transactions on Image Processing, 17(11), 2000-2014.


\bibitem{b17}
Rai, Y., Le Callet, P., \& Guillotel, P. (2017, May). Which saliency weighting for omni directional image quality assessment?. In 2017 Ninth International Conference on Quality of Multimedia Experience (QoMEX) (pp. 1-6). IEEE.

\bibitem{b18}
Xu, M., Song, Y., Wang, J., Qiao, M., Huo, L.,  Wang, Z. (2018). Predicting head movement in panoramic video: A deep reinforcement learning approach. IEEE transactions on pattern analysis and machine intelligence, 41(11), 2693-2708.

\bibitem{b19}
Sitzmann, V., Serrano, A., Pavel, A., Agrawala, M., Gutierrez, D., Masia, B., Wetzstein, G. (2018). Saliency in VR: How do people explore virtual environments?. IEEE transactions on visualization and computer graphics, 24(4), 1633-1642.


\bibitem{b20}
De Abreu, A., Ozcinar, C., \& Smolic, A. (2017, May). Look around you: saliency maps for omnidirectional images in VR applications. In 2017 Ninth International Conference on Quality of Multimedia Experience (QoMEX) (pp. 1-6). IEEE.

\bibitem{b21} 
Huang, X., Shen, C., Boix, X., \& Zhao, Q. (2015). SALICON: Reducing the semantic gap in saliency prediction by adapting deep neural networks. In Proceedings of the IEEE International Conference on Computer Vision (pp. 262-270).

\bibitem{b22} 
Lebreton, P., \& Raake, A. (2018). GBVS360, BMS360, ProSal: Extending existing saliency prediction models from 2D to omnidirectional images. Signal Processing: Image Communication, 69, 69-78.

\bibitem{b23}
Zhang, J., \& Sclaroff, S. (2013). Saliency detection: A boolean map approach. In Proceedings of the IEEE international conference on computer vision (pp. 153-160).

\bibitem{b24}
Harel, J., Koch, C.,  Perona, P. (2007). Graph-based visual saliency. In Advances in neural information processing systems (pp. 545-552).

\bibitem{b25}
T. Maugey, O. Le Meur and Z. Liu, "Saliency-based navigation in omnidirectional image," 2017 IEEE 19th International Workshop on Multimedia Signal Processing (MMSP), Luton, 2017, pp. 1-6.
\bibitem{b26}
Battisti, F., Baldoni, S., Brizzi, M.,  Carli, M. (2018). A feature-based approach for saliency estimation of omni-directional images. Signal Processing: Image Communication, 69, 53-59.

\bibitem{b27}
Fang, Y., Zhang, X., Imamoglu, N. (2018). A novel superpixel-based saliency detection model for 360-degree images. Signal Processing: Image Communication, 69, 1-7.

\bibitem{b28}
David, E. J., Gutiérrez, J., Coutrot, A., Da Silva, M. P., \& Callet, P. L. (2018, June). A dataset of head and eye movements for 360 videos. In Proceedings of the 9th ACM Multimedia Systems Conference (pp. 432-437).
ISO 690	

\bibitem{b29}
Zhang, Z., Xu, Y., Yu, J.,  Gao, S. (2018). Saliency detection in 360 videos. In Proceedings of the European Conference on Computer Vision (ECCV) (pp. 488-503).

\bibitem{b30}
Cheng, H. T., Chao, C. H., Dong, J. D., Wen, H. K., Liu, T. L., Sun, M. (2018). Cube padding for weakly-supervised saliency prediction in 360 videos. In Proceedings of the IEEE Conference on Computer Vision and Pattern Recognition (pp. 1420-1429).

\bibitem{b31}
Suzuki, T.,  Yamanaka, T. (2018, October). Saliency map estimation for omni-directional image considering prior distributions. In 2018 IEEE International Conference on Systems, Man, and Cybernetics (SMC) (pp. 2079-2084). IEEE.


\bibitem{b32}
Lebreton, P., Fremerey, S.,  Raake, A. (2018, July). V-BMS360: A video extention to the BMS360 image saliency model. In 2018 IEEE International Conference on Multimedia \&Expo Workshops (ICMEW) (pp. 1-4). IEEE.

\bibitem{b33}
Nguyen, A., Yan, Z.,  Nahrstedt, K. (2018, October). Your attention is unique: Detecting 360-degree video saliency in head-mounted display for head movement prediction. In Proceedings of the 26th ACM international conference on Multimedia (pp. 1190-1198).

\bibitem{b34}
Zhang, K.,  Chen, Z. (2018). Video saliency prediction based on spatial-temporal two-stream network. IEEE Transactions on Circuits and Systems for Video Technology, 29(12), 3544-3557.

\bibitem{b35}
Hu, H. N., Lin, Y. C., Liu, M. Y., Cheng, H. T., Chang, Y. J., \& Sun, M. (2017, July). Deep 360 pilot: Learning a deep agent for piloting through 360 sports videos. In 2017 IEEE Conference on Computer Vision and Pattern Recognition (CVPR) (pp. 1396-1405). IEEE.

\bibitem{b36}
Chao, F. Y., Zhang, L., Hamidouche, W., \& Deforges, O. (2018, July). SalGAN360: Visual saliency prediction on 360 degree images with generative adversarial networks. In 2018 IEEE International Conference on Multimedia \& Expo Workshops (ICMEW) (pp. 01-04). IEEE.

\bibitem{b37}
Qiao, M., Xu, M., Wang, Z., \& Borji, A. (2020). Viewport-dependent Saliency Prediction in 360° Video. IEEE Transactions on Multimedia.

\bibitem{b38}
Wang, F., Jiang, M., Qian, C., Yang, S., Li, C., Zhang, H., ... \& Tang, X. (2017). Residual attention network for image classification. In Proceedings of the IEEE conference on computer vision and pattern recognition (pp. 3156-3164).

\bibitem{b39}
Yang, Z., He, X., Gao, J., Deng, L., \& Smola, A. (2016). Stacked attention networks for image question answering. In Proceedings of the IEEE conference on computer vision and pattern recognition (pp. 21-29).

\bibitem{b40}
Tao, A., Sapra, K., \& Catanzaro, B. (2020). Hierarchical Multi-Scale Attention for Semantic Segmentation. arXiv preprint arXiv:2005.10821.

\bibitem{b41}
Chen, L. C., Yang, Y., Wang, J., Xu, W., \& Yuille, A. L. (2016). Attention to scale: Scale-aware semantic image segmentation. In Proceedings of the IEEE conference on computer vision and pattern recognition (pp. 3640-3649).

\bibitem{b42}
Rai, Y., Gutiérrez, J., \& Le Callet, P. (2017, June). A dataset of head and eye movements for 360 degree images. In Proceedings of the 8th ACM on Multimedia Systems Conference (pp. 205-210).

\bibitem{b43}
Sitzmann, V., Serrano, A., Pavel, A., Agrawala, M., Gutierrez, D., Masia, B., \& Wetzstein, G. (2016). How do people explore virtual environments?. arXiv preprint arXiv:1612.04335.

\bibitem{b44}
Oliva, A., \& Torralba, A. (2007). The role of context in object recognition. Trends in cognitive sciences, 11(12), 520-527.

\bibitem{b45}
He, K., Zhang, X., Ren, S., \& Sun, J. (2016). Deep residual learning for image recognition. In Proceedings of the IEEE conference on computer vision and pattern recognition (pp. 770-778).

\bibitem{b46}
Bao, Y., Zhang, T., Pande, A., Wu, H., \& Liu, X. (2017, June). Motion-prediction-based multicast for 360-degree video transmissions. In 2017 14th Annual IEEE International Conference on Sensing, Communication, and Networking (SECON) (pp. 1-9). IEEE.
\bibitem{b47}
Bylinskii, Z., Judd, T., Oliva, A., Torralba, A., \& Durand, F. (2018). What do different evaluation metrics tell us about saliency models?. IEEE transactions on pattern analysis and machine intelligence, 41(3), 740-757.

\end{thebibliography}
\end{document}